\documentclass[preprint]{elsarticle}

\usepackage{lineno,hyperref}
\modulolinenumbers[1]

\usepackage{amsmath}
\usepackage{amsfonts}
\usepackage[ruled]{algorithm2e}
\usepackage{float} 
\usepackage{array}

\journal{Journal of Computational Science}

\begin{document}


\begin{frontmatter}

\title{Generalization of Change-Point Detection in Time Series Data Based on Direct Density Ratio Estimation}

\author[mymainaddress]{Mikhail Hushchyn\corref{mycorrespondingauthor}}
\ead{hushchyn.mikhail@gmail.com}

\author[mymainaddress]{Andrey Ustyuzhanin}
\ead{andrey.u@gmail.com}

\address[mymainaddress]{National Research University Higher School of Economics, Moscow, Russia}
\cortext[mycorrespondingauthor]{Corresponding author}

\begin{abstract}
The goal of the change-point detection is to discover changes of time series distribution. One of the state of the art approaches of the change-point detection are based on direct density ratio estimation. In this work we show how existing algorithms can be generalized using various binary classification and regression models. In particular, we show that the Gradient Boosting over Decision Trees and Neural Networks can be used for this purpose. The algorithms are tested on several synthetic and real-world datasets. The results show that the proposed methods outperform classical RuLSIF algorithm. Discussion of cases where the proposed algorithms have advantages over existing methods are also provided.
\end{abstract}

\begin{keyword}
time series\sep change-point detection\sep machine learning\sep neural networks\sep density ratio 
\end{keyword}

\end{frontmatter}


\section{Introduction}
\label{Introduction}

Abrupt change of time series behaviour is called a change-point. Reasons behind these changes might be different and their detection helps to investigate them and properly react to them. For example, component failures of a system can be accompanied by change-points in time series. Detection of these changes allows to prevent the failure of the system or restore its functionality faster. Another example is signals measured in seismological stations. When seismic wave from an earthquake reaches the station it changes distribution of measured seismogram values. Detection of this change-point helps to predict coming earthquake.

Variety of change-point detection methods were proposed in recent decades and summarised in~\cite{Aminikhanghahi2017, LIU201372}. This work is focused on a group of algorithms that are based on direct density-ratio estimation. There are two main ways to estimate probability densities ratio for two samples~\cite{Sugiyama, SugiyamaMIT}. The first one requires estimation of individual densities for each of the samples. The second way estimates the density ratio directly without estimation of the individual densities. There are a range of algorithms that allow to do this. It was demonstrated in~\cite{Bickel2007} that density ratio can be estimated directly using Kernel Logistic Regression Classifier. Later several other approaches based on kernels were proposed: Kernel Mean Matching (KMM)~\cite{Huang2007}, the Kullback–Leibler importance estimation procedure (KLIEP)~\cite{kliep1}, unconstrained least-squares importance fitting (uLSIF)~\cite{ulsif1, Sugiyama2011} and extension of uLSIF called relative uLSIF (RuLSIF)~\cite{rulsif1, Yamada2013}. 

The density ratios allow to calculate of different dissimilarity scores between two samples. These scores are used in change-point detection in time series data. Application of KLIEP method for the change-point detection was demonstrated in~\cite{Kawahara2012} with Kullback–Leibler divergence as dissimilarity score.  It was shown that KLIEP outperforms kernel logistic regression classifier. Application and comparison of KLIEP, uLSIF and RuLSIF methods for change-point detection in time series are provided in~\cite{LIU201372}. According to \cite{Aminikhanghahi2017} these algorithms are considered as one of the best for change-point detection. It was demonstrated in~\cite{Nam2015} how Convolutional Neural Networks (CNN) can be trained with uLSIF loss function to detect outliers in images. Separation change point detection (SEP)~\cite{8395405} method also uses kernels with modified uLSIF loss function for change-point detection in time series.

One more interesting approach for the change analysis between two samples using decision tree and logistic regression classifiers was demonstrated in~\cite{Hido2008}. The idea of this method is that if the two samples have different distributions, a classifier will separate them and prediction accuracy will be significantly higher than 50\%. However, this method does not estimate the density ratio and it was not demonstrated how it works for change-point detection in time series.

This work describes two approaches for change-point detection in time series data based on direct density ratio estimation. The first approach uses binary classifiers for direct density ratio estimation. The second approach demonstrates how regression models can be used for the ratio estimation. In particular, there are considered algorithms based on Gradient Boosting over Decision Trees and Neural Networks.

The paper is organized as follows. The problem statement and a quality metric are given in section \ref{sec:problem}. Direct density ratio estimation algorithms and dissimilarity scores are described in section \ref{sec:scores} respectively. Section \ref{sec:cpd_algo} provides description of change-point detection algorithm. Section \ref{sec:experimetns} provides results of change-points detection in several time series. Finally, the conclusion is in section \ref{sec:conclusion}.

\section{Problem Statement}
\label{sec:problem}

Similarly to \cite{LIU201372} we define the following notations. Consider a $d$-dimensional time series that is described by a vector of observations $\textbf{x}(t) \in \mathbb{R}^{d}$ at time $t$. Let's define sequence of observations for time $t$ with length $k$ as:

\begin{equation}
\textbf{X}(t) = [\textbf{x}(t)^{T}, \textbf{x}(t-1)^{T}, ..., \textbf{x}(t-k+1)^{T}]^{T} \in \mathbb{R}^{kd}
\end{equation}

Sequence of observations is required to take into account temporal dependencies between these observations. Sample of sequences of size $n$ is defined as:

\begin{equation}
\mathcal{X}(t) = \{ \textbf{X}(t), \textbf{X}(t-1), ..., \textbf{X}(t-n+1) \}
\end{equation}

It is also implied that observation distribution changes at time $t^{*}$. The goal is to detect this change. As it will be shown in the next section, change-point detection algorithms are based on comparison of reference $\mathcal{X}_{rf}(t-n)$ and test $\mathcal{X}_{te}(t)$ samples of the time series. The idea is to estimate dissimilarity score between these two samples. The larger dissimilarity the more likely the change-point occurs at time $t-n$. Several score definitions are provided in the next section.

\begin{figure}
\centering
\includegraphics[width=1.\linewidth]{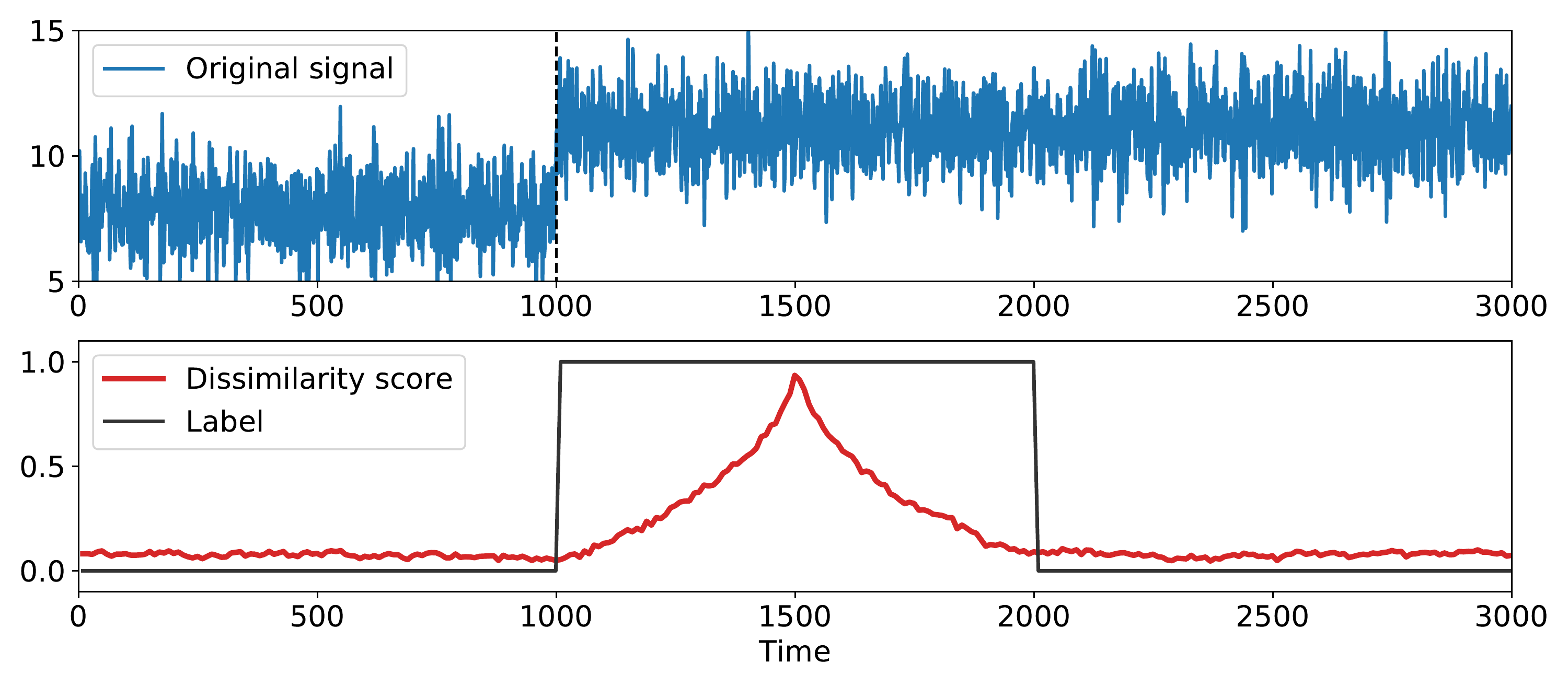}
\caption{Example of change-point detection based on direct density ratio estimation with $k=10$ and $n=500$. (Top) Time series of the original 1D signal with change-point at $t^{*}=1000$. (Bottom) Estimated dissimilarity score by the algorithm and label used for the quality metric calculation.}
\label{fig:example}
\end{figure}

Consider a time series example with a change-point at time $t^{*}=1000$ as it is shown in Fig.~\ref{fig:example}. The detection was performed with a sequence size $k=10$ and $n=500$ sequences in reference and test samples. The figure demonstrates that the dissimilarity score estimated by a change-point detection algorithm is small for times $t < t^{*}$. This corresponds to a case when reference and test samples entirely located before the change-point. Then the score starts to increase when the test sample $\mathcal{X}_{te}(t)$ contains observations after $t^{*}$ and differs from the reference sample $\mathcal{X}_{rf}(t-n)$. It reaches its maximum at $t=1500$ when the two samples are on opposite sides of the change-point. After that the score reduces until the two samples have all observation after the change-point. 

This example demonstrates that the dissimilarity score forms a peak with width of $2n$ timestamps after the change-point. Quality of the detection depends on relative height of this peak and variance of the score. The change-point is considered as detected at $\hat{t}^{*}$ when the score reaches a threshold value $D(\hat{t}^{*}) \ge \mu$. If variance of the dissimilarity score is large, the threshold value should be large enough to avoid high false alarm rate. This leads to increasing of the detection time delay $\hat{t}^{*} - t^{*}$. When the time series distribution changes not very much at $t^{*}$, the threshold might appear to be too large and the change-point will not be detected. 

To take into account these two effects the time series observation are labeled in the following way. Label 0 is attributed for all timestamps except those with $t^{*} \le t < t^{*} + 2n$ for all change-points $t^{*}$. These observations have label 1 as it is demonstrated in Fig.~\ref{fig:example}. The change-point detection quality is measured by ROC AUC score that is calculated based on estimated dissimilarity score values and defined labels.

\section{Dissimilarity Score Estimation}
\label{sec:scores}

\subsection{RuLSIF-Based Models}

Consider a regression problem to estimate the density ratio $w(X)$ directly. In the original RuLSIF~\cite{Sugiyama, ulsif1} algorithm the ratio is defined as:

\begin{equation}
    w(X) = \frac{P_{te}(X)}{(1-\alpha)P_{rf}(X) + \alpha P_{te}(X)}
\end{equation}

where $P_{te}(X)$ and $P_{rf}(X)$ are distribution densities for test $\mathcal{X}_{te}(t)$ and reference $\mathcal{X}_{rf}(t-n)$ samples respectively; $\alpha$ is an adjustable parameter. Let's $\hat{w}(X)$ is estimated ratio by a regression model. The model is fitted by minimizing the following loss function $J(X)$:

\begin{equation}
\begin{split}
J & = \frac{1}{2} \int (\hat{w}(X) - w(X))^{2} ((1-\alpha)P_{rf}(X) + \alpha P_{te}(X)) dX \\
  & = \frac{1}{2} \int \hat{w}^{2}(X) ((1-\alpha)P_{rf}(X) + \alpha P_{te}(X)) dX - \int \hat{w}(X) P_{te}(X) dX \\
  & + \frac{1}{2} \int w^{2}(X) ((1-\alpha)P_{rf}(X) + \alpha P_{te}(X)) dX
\end{split}
\end{equation}

where the last term is constant and it can be ignored during the minimization procedure. In discrete case the loss function takes the following form:

\begin{equation} \label{eq:j_rulsif}
    J = \frac{1-\alpha}{2} \frac{1}{n_{rf}} \sum_{X \in \mathcal{X}_{rf}} \hat{w}^{2}(X) + \frac{\alpha}{2} \frac{1}{n_{te}} \sum_{X \in \mathcal{X}_{te}} \hat{w}^{2}(X) - \frac{1}{n_{te}} \sum_{X \in \mathcal{X}_{te}} \hat{w}(X)
\end{equation}

where $n_{rf}$ and $n_{te}$ are numbers of sequences in reference and test samples respectively. We consider several algorithms based on this loss function.

The original RuLSIF algorithm is based on kernel methods. The density ratio $\hat{w}(X)$ is estimated as:

\begin{equation}
    \hat{w}(X) = \sum_{i=1}^{n_{kr}} \theta_{i} K(X, X_{i})
\end{equation}

where $K(X, X_{i})$ is a kernel function; $(\theta_{1}, \theta_{2}, ..., \theta_{n_{kr}})$ are parameters that are learned from the data samples. The kernel centers $\{X_{i}\}_{i=1}^{n_{kr}}$ are randomly selected from the test sample. According \cite{LIU201372} RuLSIF was used with Gaussian kernel for the change-point detection:

\begin{equation}
\label{eq:kernel}
    K(X, X_{i}) = \exp(- \frac{|| X - X_{i}||_{2}^{2}}{2\sigma^{2}})
\end{equation}

where $\sigma$ is the kernel width. It is determined using cross-validation procedure. In this work RuLSIF algorithm is taken as baseline as it was described in~\cite{LIU201372}. It was fitted with 10 gaussian kernels, $\alpha=0.1$, $\sigma = 10^{-3}, 10^{-2}, ..., 10^{3}$ and $L2$ regularization parameter values $\lambda = 10^{-3}, 10^{-2}, ..., 10^{1}$. The best values of the parameters were selected during the cross-validation procedure.

We propose to use other regression models to avoid limitations of kernel methods. In this work Gradient Boosting over Decision Trees (GBDT-RulSIF) and Neural Network (NN-RuLSIF) regression models are used for the direct density ratio estimation by minimizing the loss function in Eq.~\ref{eq:j_rulsif}. GBDT-RuLSIF algorithm is shown in Alg.~\ref{alg:gbdt_rulsif}. It uses 100 regression trees with maximum depth 6, learning rate 0.2 and $\alpha=0.1$ in the loss function. NN-RuLSIF is based on one-layer Neural Network with 10 hidden neurons with $\tanh$ activation and one output neuron with linear activation function. It was trained during 20 epochs using Adam optimizer with learning rate $\nu=0.1$, $\beta_{1}=0$, $\beta_{2}=0.9$ and batch size 32. In the loss function it is taken $\alpha=0.1$. The algorithm is described in Alg.~\ref{alg:nn_rulsif}.

\begin{algorithm}
\SetAlgoLined
\textbf{Inputs:} samples $\mathcal{X}_{rf}(t-n)$ and $\mathcal{X}_{te}(t)$ for a timestamp $t$, learning rate $\nu$, number of estimators $M$, regression trees $h_{m}(X)$, loss function $J(X)$; \\
Initialize $\hat{w}_{0}(X_{i}) = 1 + \epsilon_{i}$, $\epsilon_{i} \sim \mathcal{N}(\mu=0, \sigma=0.1)$; \\
  \For{$m = 1, ..., M$}{ 
    $\mathcal{X}_{rf}^{'}(t-n)$ $\leftarrow$ sample($\mathcal{X}_{rf}(t-n)$); \\
    $\mathcal{X}_{te}^{'}(t)$ $\leftarrow$ sample($\mathcal{X}_{te}(t)$); \\
   $z_{i} \leftarrow -\frac{\partial J(X_{i})}{\partial \hat{w}}$ = $\begin{cases}
                     -(1-\alpha)\hat{w}_{m-1}(X_{i}), & \text{if $X_{i} \in \mathcal{X}_{rf}^{'}(t-n)$}\\
                     -\alpha\hat{w}_{m-1}(X_{i}) + 1, & \text{if $X_{i} \in \mathcal{X}_{te}^{'}(t)$}
            \end{cases}$; \\
    $h_{m}(X) \leftarrow \arg\min_{h_{m}} \sum_{X_{i} \in \mathcal{X}_{rf}^{'}(t-n) \cup \mathcal{X}^{'}_{te}(t)} (h_{m}(X_{i}) - z_{i})^{2}$; \\
    $\hat{w}_{m}(X) \leftarrow \hat{w}_{m-1}(X) + \nu h_{m}(X)$; \\
   }
 \Return Approximation function $ \hat{w}_{M}(X) = \hat{w}_{0}(X) + \sum_{m=1}^{M}\nu h_{m}(X) $
 \caption{Fit GBDT-RuLSIF algorithm.}
 \label{alg:gbdt_rulsif}
\end{algorithm}

\begin{algorithm}
\SetAlgoLined
\textbf{Inputs:} samples $\mathcal{X}_{rf}(t-n)$ and $\mathcal{X}_{te}(t)$, learning rate $\nu$, neural network weights $\theta$, number of epochs $M$, Adam hyperparameters $\nu, \beta_{1}, \beta_{2}$; \\
Initialize parameters $\theta_{0}$ of a neural network $\hat{w}(X, \theta_{0})$; \\
  \For{$m = 1, ..., M$}{ 
    \For{each batch}{
    $\mathcal{X}_{rf}^{'}(t-n)$ $\leftarrow$ sample batch from $\mathcal{X}_{rf}(t-n)$; \\
    $\mathcal{X}_{te}^{'}(t)$ $\leftarrow$ sample batch from $\mathcal{X}_{te}(t)$; \\
    $J(\theta) \leftarrow \frac{1-\alpha}{2n_{rf}} \sum_{X \in \mathcal{X}^{'}_{rf}(t-n)} \hat{w}^{2}(X, \theta) + \frac{\alpha}{2n_{te}} \sum_{X \in \mathcal{X}_{te}^{'}(t))} \hat{w}^{2}(X, \theta)$; \\
    ~~~~~~~~~~$- \frac{1}{n_{te}} \sum_{X \in \mathcal{X}_{te}^{'}(t)} \hat{w}(X, \theta)$; \\
    $\theta \leftarrow \text{Adam}(\nabla_{\theta}J(\theta), \theta, \nu, \beta_{1}, \beta_{2})$;\\
    }
   }
 \Return Approximation function $\hat{w}(X, \theta)$
 \caption{Fit NN-RuLSIF algorithm.}
 \label{alg:nn_rulsif}
\end{algorithm}

Estimated density ratios $\hat{w}(X)$ are used to calculate dissimilarity between the reference and test samples. For the RuLSIF-based models the following dissimilarity score is used:

\begin{equation} \label{eq:d_reg}
    D = PE(P_{te}|(1-\alpha)P_{rf} + \alpha P_{te}) + PE(P_{rf}|(1-\alpha)P_{te} + \alpha P_{rf})
\end{equation}

where $PE(P|Q)$ is the Pearson $\chi^{2}-$divergence that defined as:

\begin{equation}
    PE(P||Q) = \int Q(x) (\frac{P(x)}{Q(x)} - 1)^{2} dx
\end{equation}

Rewrite this equation for discrete case:

\begin{equation}
\begin{split}
    PE(P||Q) = \sum_{x \sim P(x)} \frac{P(x)}{Q(x)} - 1
\end{split}
\end{equation}

Then, the dissimilarity score takes the following form:

\begin{equation} \label{eq:d_reg}
\begin{split}
D & = \frac{1}{n_{te}} \sum_{X \in \mathcal{X}_{te}} \frac{P_{te}(X)}{(1-\alpha)P_{rf}(X) + \alpha P_{te}(X)} \\
  & + \frac{1}{n_{rf}} \sum_{X \in \mathcal{X}_{rf}} \frac{P_{rf}(X)}{(1-\alpha)P_{te}(X) + \alpha P_{rf}(X)} - 2 \\
  & = \frac{1}{n_{te}} \sum_{X \in \mathcal{X}_{te}} w(X) + \frac{1}{n_{rf}} \sum_{X \in \mathcal{X}_{rf}} w'(X) - 2
\end{split}
\end{equation}

where $w(X)$ is estimated by the RuLSIF-based models using $\mathcal{X}(t-n)$ and $\mathcal{X}(t)$ as reference and test samples respectively; $w'(X)$ is estimated by the models using $\mathcal{X}(t-n)$ and $\mathcal{X}(t)$ as test and reference samples respectively. So, in this case the models are fitted twice.

\subsection{Binary Classifiers}

In additional to RuLSIF-based methods we demonstrate how binary classifiers can be used for the change-point detection in time series. Suppose the reference sample $\mathcal{X}_{rf}(t-n)$ has label $y=0$ and the test sample $\mathcal{X}_{te}(t)$ has label $y=1$. Also suppose that a probabilistic binary classifier is used to separate these two samples. The classifier output is denoted as $P(y=1|X)$ and has the meaning of probability that a sequence of observations $X$ belongs to the test sample. The distribution density ratio $w(X)$ is defined as:

\begin{equation}
    w(X) = \frac{P_{te}(X)}{P_{rf}(X)}
\end{equation}

where $P_{te}(X)$ and $P_{rf}(X)$ are distribution densities for test and reference samples respectively. In terms of the probabilistic classifier $P_{te}(X) = P(X|y=1)$ and $P_{rf}(X) = P(X|y=0)$. Using Bayes' theorem let's write out the following expressions:

\begin{equation}
    P(X|y=1) = \frac{P(y=1|X)P(X)}{P(y=1)}
\end{equation}

\begin{equation}
    P(X|y=0) = \frac{P(y=0|X)P(X)}{P(y=0)}
\end{equation}

Then the density ratio is defined as follows:

\begin{equation}
    w(X) = \frac{P(y=0)}{P(y=1)} \frac{P(y=1|X)}{P(y=0|X)}
\end{equation}

where $P(y=0)$ and $P(y=1)$ are estimated as:

\begin{equation}
    P(y=0) = \frac{N_{0}}{N_{0}+N_{1}}
\end{equation}

\begin{equation}
    P(y=1) = \frac{N_{1}}{N_{0}+N_{1}}
\end{equation}

where $N_{0}$ and $N_{1}$ are sizes of reference and test samples respectively. In this work it is supposed that the samples have the same sizes. Also, from the probabilistic classifier we know that $P(y=0|X) = 1 - P(y=1|X)$. Taking into account these two properties we write out the estimated density ratio $\hat{w}(X)$ expression as:

\begin{equation} \label{eq:w_clf}
    \hat{w}(X) = \frac{P(y=1|X)}{1 - P(y=1|X)} = \frac{f(X)}{1 - f(X)}
\end{equation}

where $f(X) = P(y=1|X)$ for simplicity. Different binary classifiers can be used instead of probabilistic one. In this case density ratios are estimated with some inaccuracy. However, it is enough for change-point detection in time series.

To estimate dissimilarity between the reference and test samples in case of using binary classifiers the following score is used:

\begin{equation} \label{eq:d_clf}
    D = KL(P_{te}||P_{rf}) + KL(P_{rf}||P_{te})
\end{equation}

where $KL(P|Q)$ is the Kullback–Leibler divergence that defined as:

\begin{equation}
    KL(P||Q) = \int P(x) \log \frac{P(x)}{Q(x)} dx
\end{equation}

where $P$ and $Q$ are probability distributions. Using this definition and Eq.~\ref{eq:w_clf} rewrite the score $D$ for discrete case:

\begin{equation} \label{eq:d_clf}
\begin{split}
    D & = \int P_{te}(X) \log \frac{P_{te}(X)}{P_{rf}(X)} dx +          \int P_{rf}(X) \log \frac{P_{rf}(X)}{P_{te}(X)} dx \\
      & = \frac{1}{n_{te}} \sum_{X \in \mathcal{X}_{te}} \log \frac{P_{te}(X)}{P_{rf}(X)} +       \frac{1}{n_{rf}} \sum_{X \in \mathcal{X}_{rf}} \log \frac{P_{rf}(X)}{P_{te}(X)} \\
      & = \frac{1}{n_{te}} \sum_{X \in \mathcal{X}_{te}} \log \frac{f(X)}{1 - f(X)} +  \frac{1}{n_{rf}} \sum_{X \in \mathcal{X}_{rf}} \log \frac{1 - f(X)}{f(X)}
\end{split}
\end{equation}

Scores based on other $f$-divergences also can be used. Conventional classification metrics like ROC AUC or accuracy are also suitable for the dissimilarity measures. However, our experiments show that the proposed score in Eq.~\ref{eq:d_clf} is better for the change-point detection problem in time series using binary classifiers.

In this work the following classifiers are used: Gradient Boosting over Decision Trees (GBDT) implemented in XGBoost python library and Neural Network (NN) using PyTorch library. GBDT model was fitted using 100 decision trees with maximum depth 6, learning rate 0.1 and binary cross-entropy loss function. NN model uses one-layer Neural Network with 10 hidden neurons with $\tanh$ activation and one output neuron with sigmoid activation function. It was fitted with binary cross-entropy loss function during 20 epochs using Adam optimizer with learning rate $\nu=0.1$, $\beta_{1}=0$, $\beta_{2}=0.9$ and batch size 32.

\section{Change-Point Detection Algorithm}
\label{sec:cpd_algo}

\begin{algorithm}
\SetAlgoLined
\textbf{Inputs:} time series $\{X(t)\}_{t=k}^{T}$; $k$ - size of a sequence $X(t)$; $n$ - size of sample of sequences $\mathcal{X}(t)$; $m$ - number of iterations; model - direct density ratio estimation model; $W(t)$ - vector of density ratios $\hat{w}(X)$ for $\mathcal{X}(t)$; $D(t)$ - dissimilarity score; \\
\textbf{Initialization:} $t \leftarrow k + 2n$; \\
\While{$t \le T$}{
  create samples $\mathcal{X}_{rf}(t-n)$ and $\mathcal{X}_{te}(t)$; \\
  $D(t) = 0$;\\
  $i = 1$; \\
  \For{$i = 1, ..., m$}{ 
    $\mathcal{X}_{rf}^{train}(t-n)$, $\mathcal{X}_{rf}^{valid}(t-n)$ $\leftarrow$ random\_split($\mathcal{X}_{rf}(t-n)$, size=0.5); \\
    $\mathcal{X}_{te}^{train}(t)$, $\mathcal{X}_{te}^{valid}(t)$ $\leftarrow$ random\_split($\mathcal{X}_{te}(t)$, size=0.5); \\
    model.fit($\mathcal{X}_{rf}^{train}(t-n)$, $\mathcal{X}_{te}^{train}(t)$);\\
    $W_{rf}^{valid}(t-n)$ $\leftarrow$ model.predict($\mathcal{X}_{rf}^{valid}(t-n)$);\\
    $W_{te}^{valid}(t)$ $\leftarrow$ model.predict($\mathcal{X}_{te}^{valid}(t)$);\\
    $D(t) \leftarrow D(t)$ + $\frac{1}{m}$ dissimilarity($W_{rf}^{valid}(t-n)$, $W_{te}^{valid}(t)$);\\
   }
 $t \leftarrow t + \delta t$; \\
 }
 \Return{$\{D(t)\}_{t=k}^{T}$} 
 \caption{Change-point detection algorithm.}
 \label{alg:cpd}
\end{algorithm}

Change-point detection algorithm is based on comparison of reference $\mathcal{X}_{rf}(t-n)$ and test $\mathcal{X}_{te}(t)$ samples in two consecutive time windows. The RuLSIF-based models or binary classifiers are used to estimate dissimilarity score $D(t)$ between the samples for the timestamp $t$. To do this the samples are splitted into train and validation subsamples. The train subsample is used to fit the models as it was described in Sec.~\ref{sec:scores}. Then, these models are used to estimate the density ratios $\hat{w}(X)$ on the validation subsample and to calculate the dissimilarity score $D(t)$. This procedure is repeated for all timestamps with time step $\delta t$. If reference and test samples have the same distribution of observations the dissimilarity score will be close to 0. Larger difference between the distributions higher the dissimilarity score. Change-point is considered as detected at time $\hat{t}^{*}$ when $D(\hat{t}^{*}) \ge \mu$, where $\mu$ is a threshold value. The change-point detection algorithm is described in Alg.~\ref{alg:cpd}. 

For all experiments presented in this work the following parameter values are used: $k=10$, $n=500$, $m=1$. For experiments with the Kepler data $n=200$ was used due to small time intervals between two consecutive change points.

\section{Experiments}
\label{sec:experimetns}

\subsection{Synthetic datasets}

\begin{figure}[t]
\centering
\includegraphics[width=1.\linewidth]{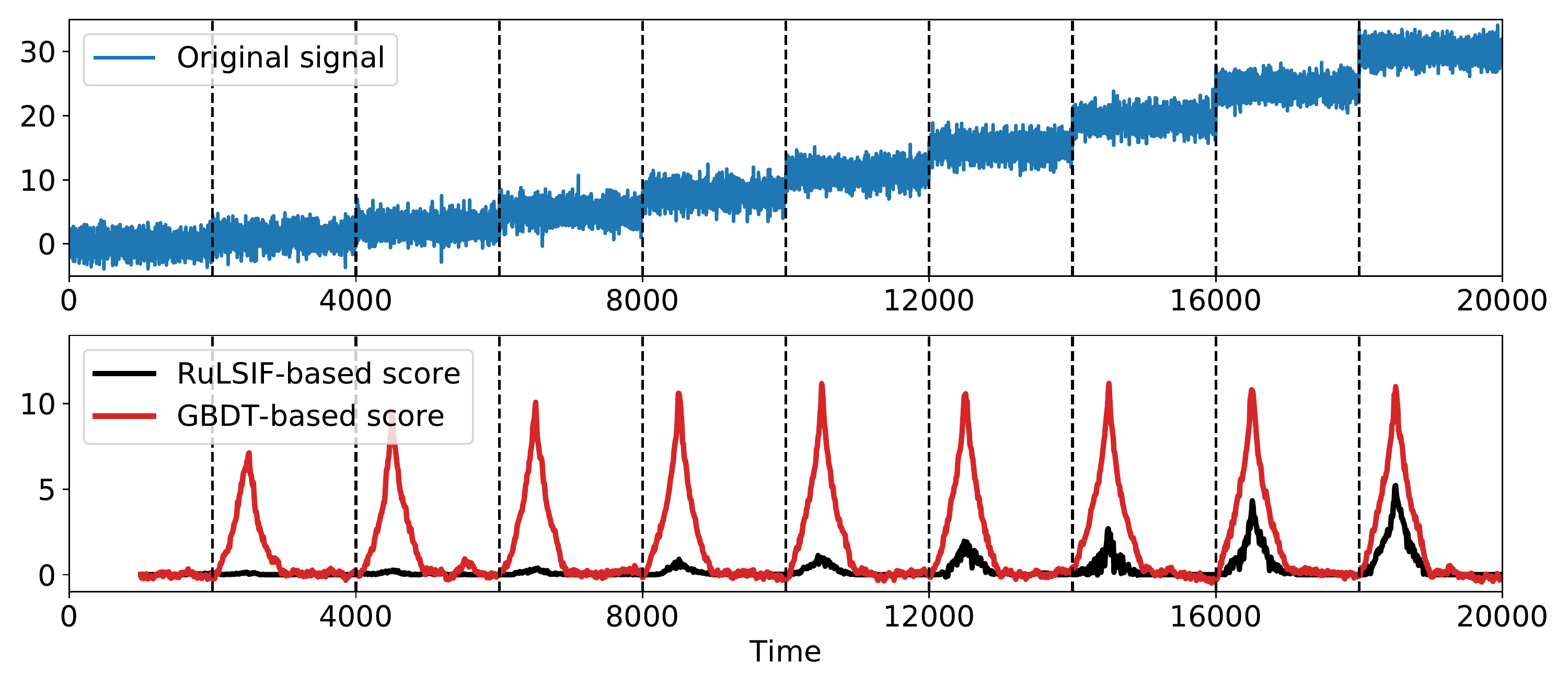}
\caption{Demonstration of the change-point detection algorithms performance on dataset 1. (Top) The first component of the time series. (Bottom) Results of the detection using RuLSIF and GBDT binary classifier.}
\label{fig:dataset1}
\end{figure}

Similar to \cite{LIU201372} the following synthetic datasets are used to demonstrate and compare change-point detection algorithms:

\begin{description}

\item[Dataset 1:] The first dataset consists of two components. The first component is generated using 1-dimensional auto-regressive process:

\begin{equation}
x_{1}(t) = 0.6 x_{1}(t-1) - 0.5 x_{1}(t-2) + \epsilon_{1}(t)
\end{equation}

where $\epsilon_{1}(t) \sim \mathcal{N}(\mu, \sigma)$ is Gaussian noise with mean $\mu$ and standard deviation $\sigma=1$. Change-points are generated every 2000 timestamps by changing mean $\mu$ in the following way:

\begin{equation}
\mu_{N} = 
\begin{cases} 
0, & \mbox{if } N = 1 \\ 
\mu_{N-1} + 0.5N, & \mbox{if } N = 2, ..., 10 
\end{cases}
\end{equation}

where $N$ is an integer that estimated as $2000(N-1) + 1 \le t \le 2000N$. The second component is Gaussian noise that is generated as following:

\begin{equation}
x_{2}(t) = \epsilon_{2}(t)
\end{equation}

where $\epsilon_{2}(t) \sim \mathcal{N}(\mu, \sigma)$ is Gaussian noise with mean $\mu=0$ and standard deviation $\sigma=5$. This component is added to demonstrate that non-kernels methods are less sensitive to uninformative input features.

\item[Dataset 2:] The second dataset is generated using the same auto-regressive process as for Dataset 1, but change-points are generated by changing standard deviation $\sigma$ of the first component with constant mean $\mu=0$: 

\begin{equation}
\sigma_{N} = 
\begin{cases} 
1, & \mbox{if } N = 1 \\ 
1 + 0.25N, & \mbox{if } N = 2, ..., 10 
\end{cases}
\end{equation}

where $N$ is an integer that estimated as $2000(N-1) + 1 \le t \le 2000N$.

\item[Dataset 3:] The third dataset is a periodic 1-dimensional time series that generated as:

\begin{equation}
x(t) = \sin(\omega t) + \epsilon (t)
\end{equation}

where $\epsilon(t) \sim \mathcal{N}(\mu, \sigma)$ is Gaussian noise with mean $\mu=0.5$ and standard deviation $\sigma=1$. Change-points are generated every 2000 time steps by changing frequency $\omega$ in the following way:

\begin{equation}
\omega_{N} = 
\begin{cases} 
1, & \mbox{if } N = 1 \\ 
\omega \log(e + 0.5 N), & \mbox{if } N = 2, ..., 10 
\end{cases}
\end{equation}

where $N$ is an integer that estimated as $2000(N-1) + 1 \le t \le 2000N$.

\end{description}

\begin{figure}[t]
\centering
\includegraphics[width=1.\linewidth]{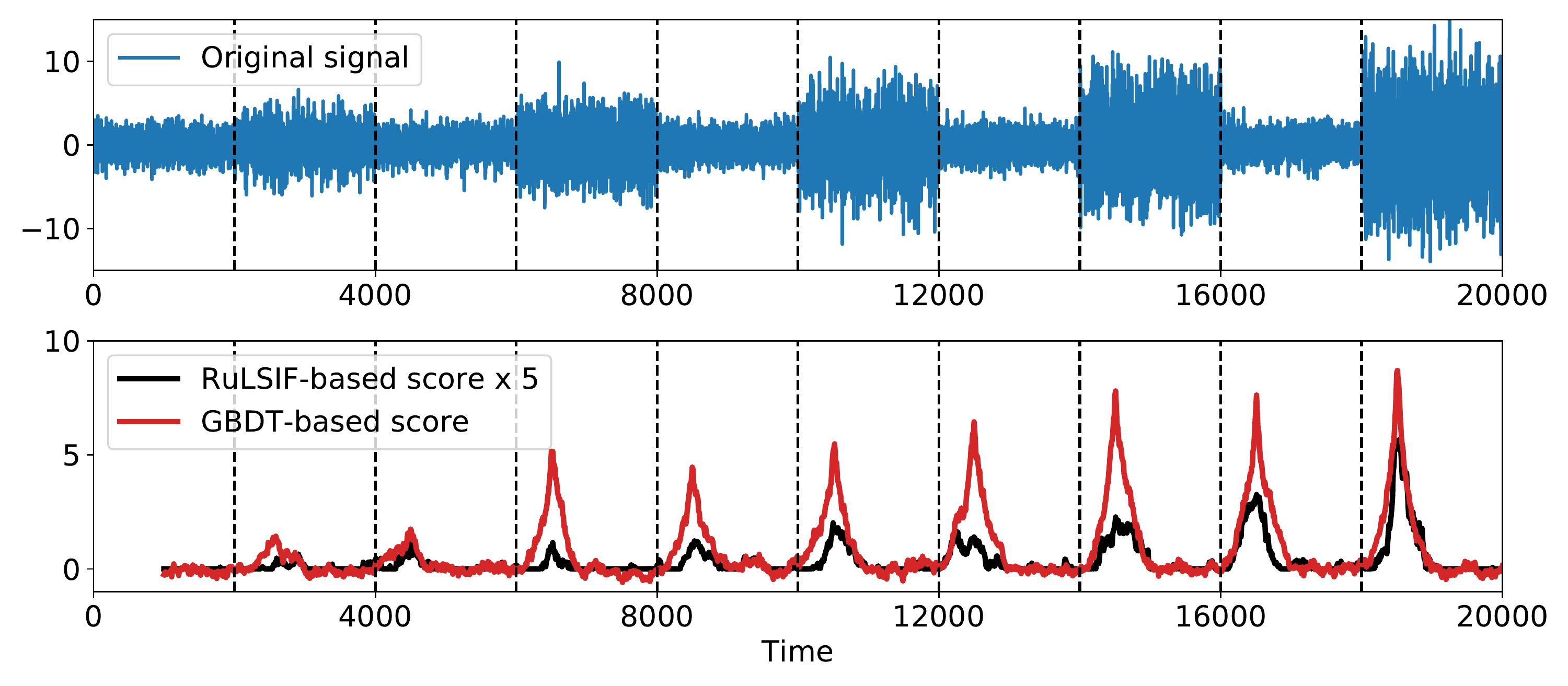}
\caption{Demonstration of the change-point detection algorithms performance on dataset 2. (Top) The first component of the time series. (Bottom) Results of the detection using RuLSIF and GBDT binary classifier.}
\label{fig:dataset2}
\end{figure}

\begin{figure}[t]
\centering
\includegraphics[width=1.\linewidth]{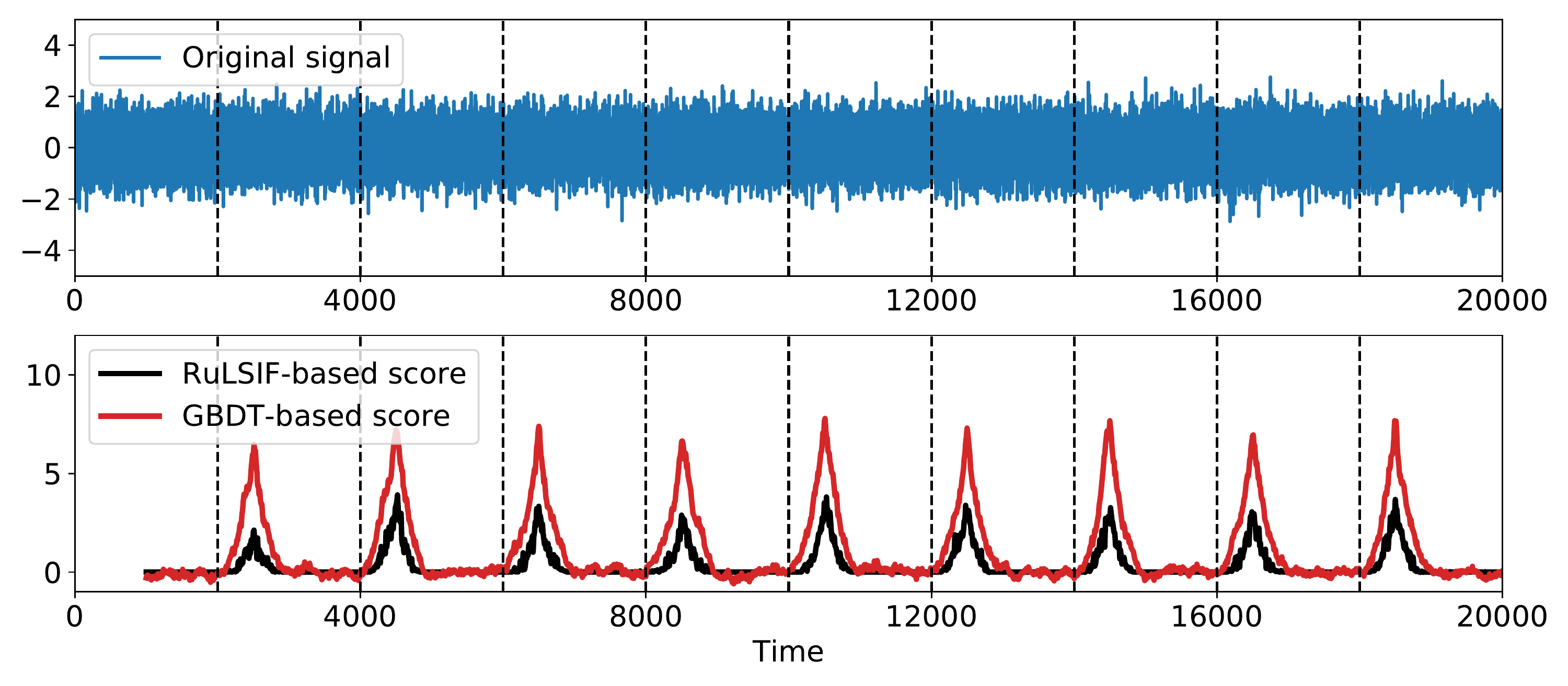}
\caption{Demonstration of the change-point detection algorithms performance on dataset 3. (Top) The first component of the time series. (Bottom) Results of the detection using RuLSIF and GBDT binary classifier.}
\label{fig:dataset3}
\end{figure}

Each dataset was generated 10 times and the algorithms results were averaged. Demonstrations of change-point detection using RuLSIF and GBDT binary classifier for the synthetic datasets are provided in Fig.~\ref{fig:dataset1}-\ref{fig:dataset3}. ROC AUC values for all algorithms are presented in Tab.~\ref{tab:synthetic}. The table shows that all of them work well and all new considered algorithms outperform RuLSIF. This is explained by the fact that RuLSIF is based on kernel methods and uses distances between observations as it is show in Eq.~\ref{eq:kernel}. Time series in datasets 1 and 2 contain the noise component that does not provide any information about change-points. This component decreases the dissimilarity score estimated by RuLSIF and makes the change-point detection harder. Algorithms based on Gradient Boosting and Neural Networks classifiers are able to recognize uninformative components and decrease their influence on the dissimilarity score. Moreover, they can estimate importance of all components and use it for better change-point detection.

\begin{table}
\centering
\begin{tabular}{|p{3cm}|p{2.3cm}|p{2.3cm}|p{2.3cm}|}
\hline
Algorithm &  Dataset 1 &  Dataset 2 &  Dataset 3 \\
\hline
RuLSIF &        0.867 $\pm$ 0.003 &  0.760 $\pm$ 0.003 &  0.843$ \pm$ 0.003 \\
GBDT-RuLSIF &   0.954 $\pm$ 0.002 &  0.892 $\pm$ 0.003 &  0.919$ \pm$ 0.002 \\
NN-RuLSIF &     0.950 $\pm$ 0.002 &  0.833 $\pm$ 0.003 &  0.941$ \pm$ 0.002 \\
NN &            0.951 $\pm$ 0.001 &  0.816 $\pm$ 0.003 &  0.933$ \pm$ 0.003 \\
GBDT &          0.960 $\pm$ 0.001 &  0.895 $\pm$ 0.002 &  0.930$ \pm$ 0.002 \\
\hline
\end{tabular}
\caption{ROC AUC values for change-point detection algorithms and for the synthetic datasets.}
\label{tab:synthetic}
\end{table}

In general, we recommend to use change-point detection algorithms based on NN or GBDT binary classifiers and regression models with RuLSIF loss function instead of kernel methods in case of high dimensional time series. They have better results when not all time series components describe change-points, less sensitive to uninformative components and are able to recognize more complicated changes of time series distribution.

\subsection{Real-world datasets}

\renewcommand{\thefootnote}{\arabic{footnote}}

\begin{figure}[t]
\centering
\includegraphics[width=1.\linewidth]{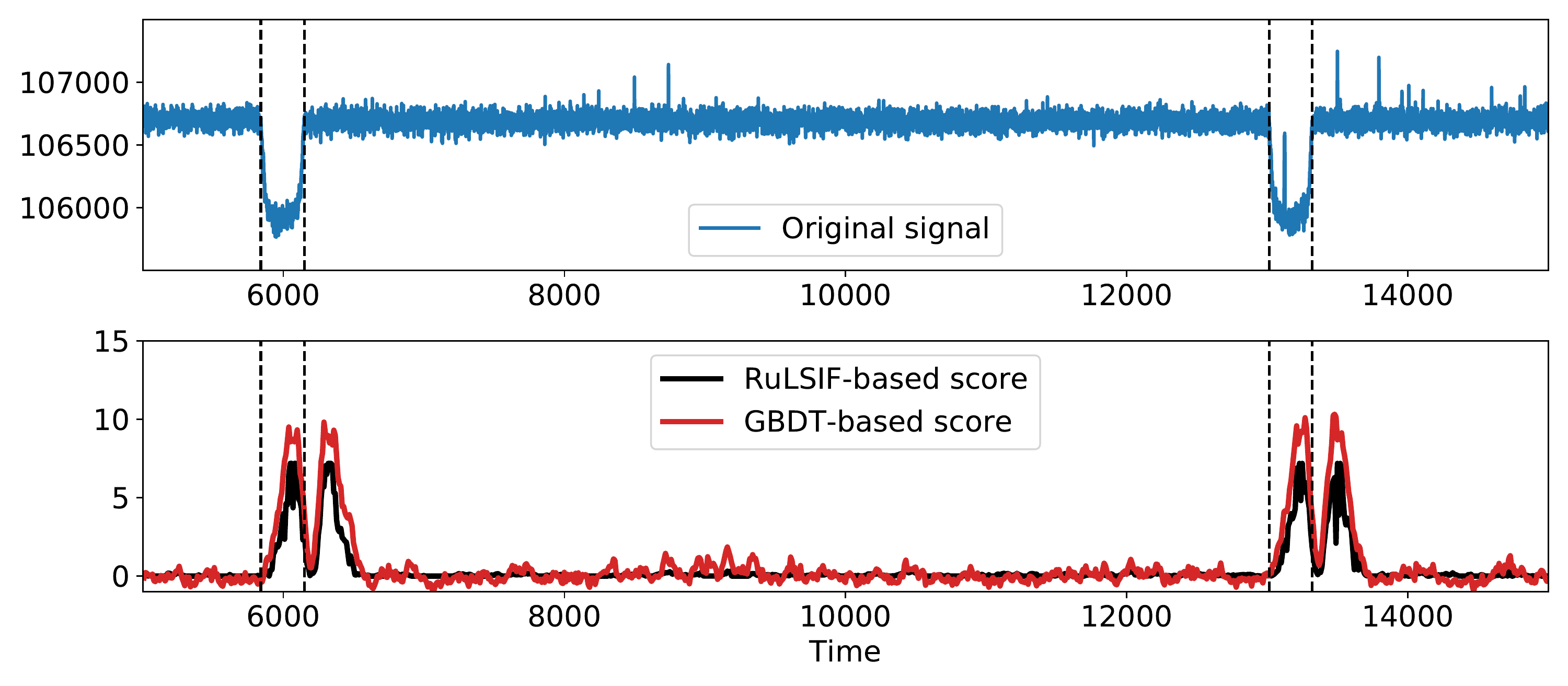}
\caption{(Top) The Kepler light curve for a star with an exoplanet. (Bottom) Dissimilarity score estimated by change-point detection algorithm based on RuLSIF and GBDT classifier.}
\label{fig:kepler}
\end{figure}

To demonstrate work of the change-point detection algorithms on real-world data the two following datasets are used:

\begin{description}

\item[The Kepler data:] Data from the Kepler spacecraft that was launched in March 2009. Its mission was to search for transit-driven exoplanet located within the habitable zones of Sun-like stars. In this work we use the Kepler light curves~\cite{kepler2019} with Data Conditioning Simple Aperture Photometry (DCSAP) data for 10 stars with exoplanets. 

\item[IRIS data:] The second dataset consists of data from Incorporated Research Institutions for Seismology (IRIS). For the demonstration we took seismograms~\cite{iris2019} recorded on seismological stations over the world for one earthquake.


\end{description}

Example of the Kepler light curve is demonstrated in Fig.~\ref{fig:kepler}. The spacecraft measured light flux coming from a star. When an exoplanet goes between the star and the spacecraft the light flux drops as it is shown in the figure. Parameters of this drop allows to estimate properties of the exoplanet. One light curve has about 10 such drops and, as result, about 20 change-points. Experiments were done on 10 light curves of different stars and results were averaged. Demonstrations of the change-point detection based on RuLSIF and GBDT classifier are shown in Fig.~\ref{fig:kepler}. ROC AUC values for the all algorithms are provided in Tab.~\ref{tab:real}.

\begin{figure}
\centering
\includegraphics[width=1.\linewidth]{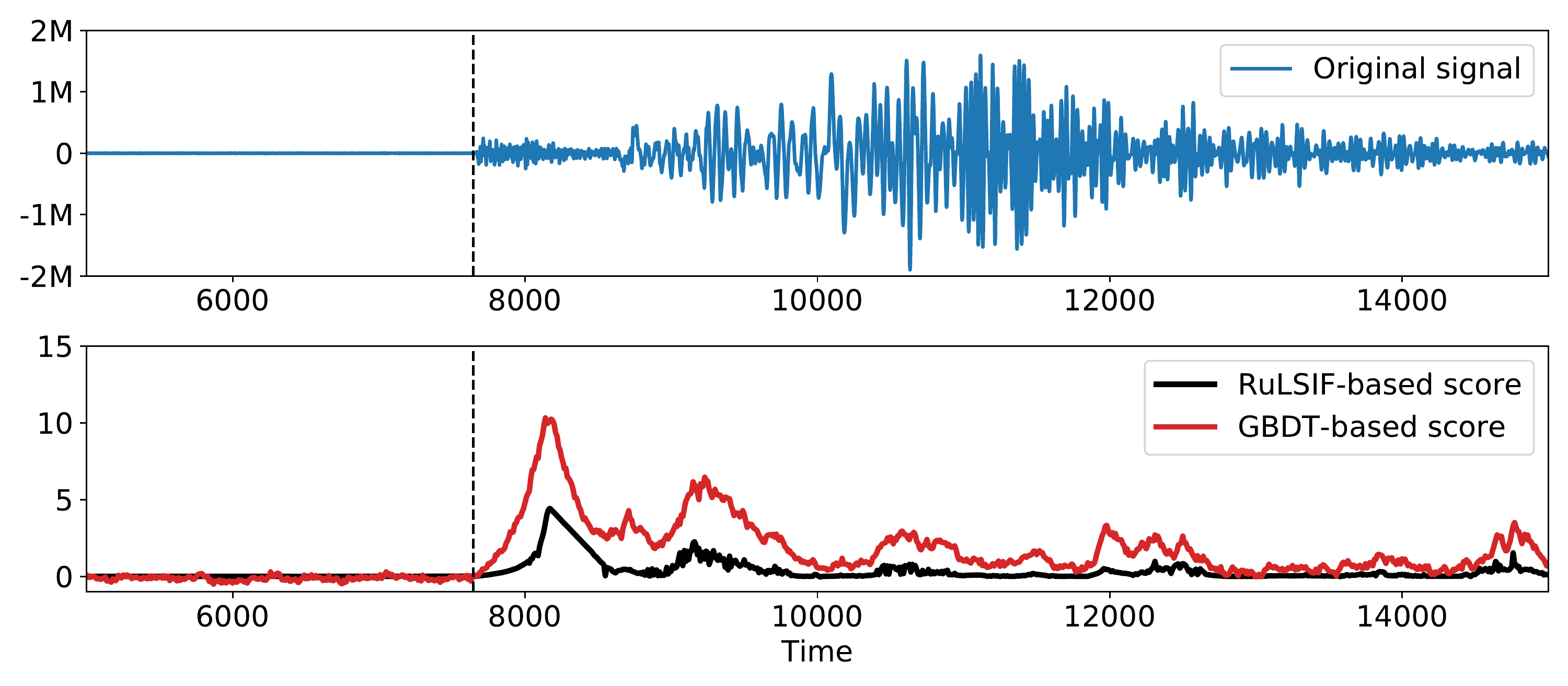}
\caption{(Top) IRIS seismogram from one station. (Bottom) Dissimilarity score estimated by change-point detection algorithm based on RuLSIF and GBDT classifier.}
\label{fig:iris}
\end{figure}

Example of a signal recorded by a seismological station is show in Fig.~\ref{fig:iris}. When seismic wave reaches the station the signal variance increases and changes with time. Before an earthquake detected dissimilarity score takes small values. The score increases and forms the first peak right after the seismic wave registration. Then the score changes with changing of the signal. It has trend to slow down with the seismic wave attenuation as it is demonstrated in the figure. ROC AUC values for the all algorithms are provided in Tab.~\ref{tab:real}.

\begin{table}[t]
\centering
\begin{tabular}{|p{3cm}|p{3cm}|p{3cm}|}
\hline
Algorithm &  Kepler data &  IRIS data \\
\hline
RuLSIF &        0.844 $\pm$ 0.005 &  0.971 $\pm$ 0.003 \\
GBDT-RuLSIF &   0.940 $\pm$ 0.002 &  0.978 $\pm$ 0.004 \\
NN-RuLSIF &     0.825 $\pm$ 0.004 &  0.942 $\pm$ 0.002 \\
NN &            0.943 $\pm$ 0.002 &  0.971 $\pm$ 0.003 \\
GBDT &          0.950 $\pm$ 0.002 &  0.980 $\pm$ 0.002 \\
\hline
\end{tabular}
\caption{ROC AUC values for change-point detection algorithms and for the real world datasets.}
\label{tab:real}
\end{table}

\section{Conclusion}
\label{sec:conclusion}

Two different approaches for change-point detection in time series based on direct density-ratio estimation are presented in the work. It was demonstrated that this problem can be solved using different binary classification and regression algorithms in machine learning, but not only based on kernel methods. In particular, change-point detection algorithms based on Gradient Boosting over Decision Trees and Neural Networks classification and regression were discussed in this work. It was shown that the proposed algorithms outperform the classical RuLSIF on a set of synthetic and real-world datasets.

\section{Acknowledgments}

We wish to thank Denis Derkach and Vladislav Belavin for their useful comments and discussions of this work.

The research was carried out with the financial support of the Ministry of Science and Higher Education of Russian Federation within the framework of the Federal Target Program “Research and Development in Priority Areas of the Development of the Scientific and Technological Complex of Russia for 2014-2020”. Unique identifier – RFMEFI58117X0023, agreement 14.581.21.0023 on 03.10.2017.

\bibliography{elsarticle-template}

\end{document}